\documentclass{article}
\usepackage[utf8]{inputenc}
\usepackage[final]{neurips_rl2019}
\usepackage{graphicx}
\usepackage{color}
\usepackage[normalem]{ulem}
\usepackage{url}

\usepackage{graphicx}
\usepackage{subcaption}
\usepackage{amsmath}
\usepackage{nicefrac}
\usepackage{natbib}
\usepackage{graphicx}
\usepackage{subcaption}
\usepackage{balance}
\usepackage{float}
\usepackage{algorithm}
\usepackage{algorithmic}

\usepackage[colorinlistoftodos]{todonotes}
\usepackage[T1]{fontenc}
\usepackage{booktabs}       
\usepackage{amsfonts}       
\usepackage{microtype}      
\usepackage{float}

\usepackage{hyperref}

\usepackage{url}

\usepackage[english]{babel}
\usepackage[utf8]{inputenc}
\usepackage{amssymb}
\usepackage{amsthm}
\usepackage{wrapfig}

\usepackage{color}

\title{Augmenting learning using symmetry in a biologically-inspired domain}
\author{Shruti Mishra$^{1,2}$, Abbas Abdolmaleki$^1$, Arthur Guez$^1$, Piotr Trochim$^1$, Doina Precup$^{1,3}$\\
$^1$ DeepMind\\
$^2$ Harvard University\\
$^3$ McGill University
}
\date{August 2019}

\begin{document}

\maketitle

\begin{abstract}
Invariances to translation, rotation and other spatial transformations are a hallmark of the laws of motion, and have widespread use in the natural sciences to reduce the dimensionality of systems of equations e.g.\ \cite{wilson1971renormalization, barenblatt1996scaling}. 
In supervised learning, such as in image classification tasks, rotation, translation and scale invariances are used to augment training datasets, such as in \cite{krizhevsky2012imagenet}. 
In this work, we use data augmentation in a similar way, exploiting symmetry in the quadruped domain of the DeepMind control suite \cite{deepmindcontrolsuite2018} to add to the trajectories experienced by the actor in the actor-critic algorithm of Abdolmaleki \emph{et al.}\ \cite{abdolmaleki2018maximum}. In a data-limited regime, the agent using a set of experiences augmented through symmetry is able to learn faster. Our approach can be used to inject knowledge of invariances in the domain and task to augment learning in robots, and more generally, to speed up learning in realistic robotics applications. 
\end{abstract}

\section{Introduction}
Reinforcement learning (RL) agents are able to perform locomotion tasks using continuous actions and observations in animal-like domains such as the planar walker, half-cheetah, and humanoid \cite{lillicrap2015continuous,haarnoja2018soft}. While the agents learn policies that successfully maximise a cumulative return, the behaviours of agents in such domains often appear idiosyncratic to humans in at least two ways. Most notably, the locomotion behaviour in animals is associated with a natural rhythm and stereotypy, such as in the case of a walking dog, galloping horse, or the tripod gait of an ant. In the absence of this spatial order and temporal rhythm in simulated agents, the behaviours of the learned policies can look unnatural.

 Idiosyncratic behaviour can arise in simulated agents because the physical simulation is incorrect, the task does not capture the tasks executed by embodied agents, or the constraints on a simulated agent are different from those on an embodied agent. The physical simulation itself is an inaccurate and incomplete model of the physical world -- it allows perfect floors, instantaneous, noise-free actuation, and a windless environment. Embodied agents necessarily have behaviour policies that are robust to deviations from any precise environment crafted for simulation. Additionally, the mechanical and biological constraints that restrict the behaviour of an embodied agent, a robot or an animal, are different from those experienced by simulated agents. Embodied agents are also necessarily affected by notions of mechanical stress and fatigue. For example, an embodied walker would be less likely to drag a leg, due to the notion of wear, or flail around its arms, due to considerations of energy and efficiency.

In this work, we leverage order in the spatial structure of the domain. Specifically, the vast majority of animals have at least one plane of external symmetry, which is reflected in their stereotyped gaits when they are walking or running. Similarly, for simulated bodies with nearly identical limbs, if the agent knows how to move one limb relative to the body in a particular way, it can do the same for the other similar limbs. We use symmetry as a natural inductive bias in order to reduce the space of behaviour policies that can be achieved by agents, as well as to explore if this approach can boost the learning process.

\section{Related work} 
We try to use invariances in the physical domain/task to augment learning. Invariances to scale, orientation, translation and relative motion are common to the natural world and a hallmark of the dynamics of tangible objects and of continuous fields. The use of such invariances to reduce the dimensionality of the equations governing a physical process is thus ubiquitous in the natural sciences \cite{barenblatt1996scaling, wilson1971renormalization}. In machine learning, supervised classification tasks use rotation, symmetry and scale invariances to augment training data to get improved performance \cite{baird1992document,  simard2003best, krizhevsky2012imagenet}. In RL, symmetry and rotation invariances have been used to augment data in AlphaGo \cite{silver2016mastering}.
In robotics and continuous control applications, some ways to improve the real-world suitability and sample efficiency of RL agents include curriculum learning \cite{bengio2009curriculum}, learning from motion-capture data \cite{merel2018neural}, and  using constraint-based objective functions \cite{bohez2019value}. Our work follows the same theme of making the policies learned by RL agents more realistic, and aiming to make the learning process efficient. 

\section{Domain and tasks}
\label{sec:Method}
\begin{wrapfigure}{r}{0.5\textwidth}
\centering
\includegraphics[width=1\linewidth]{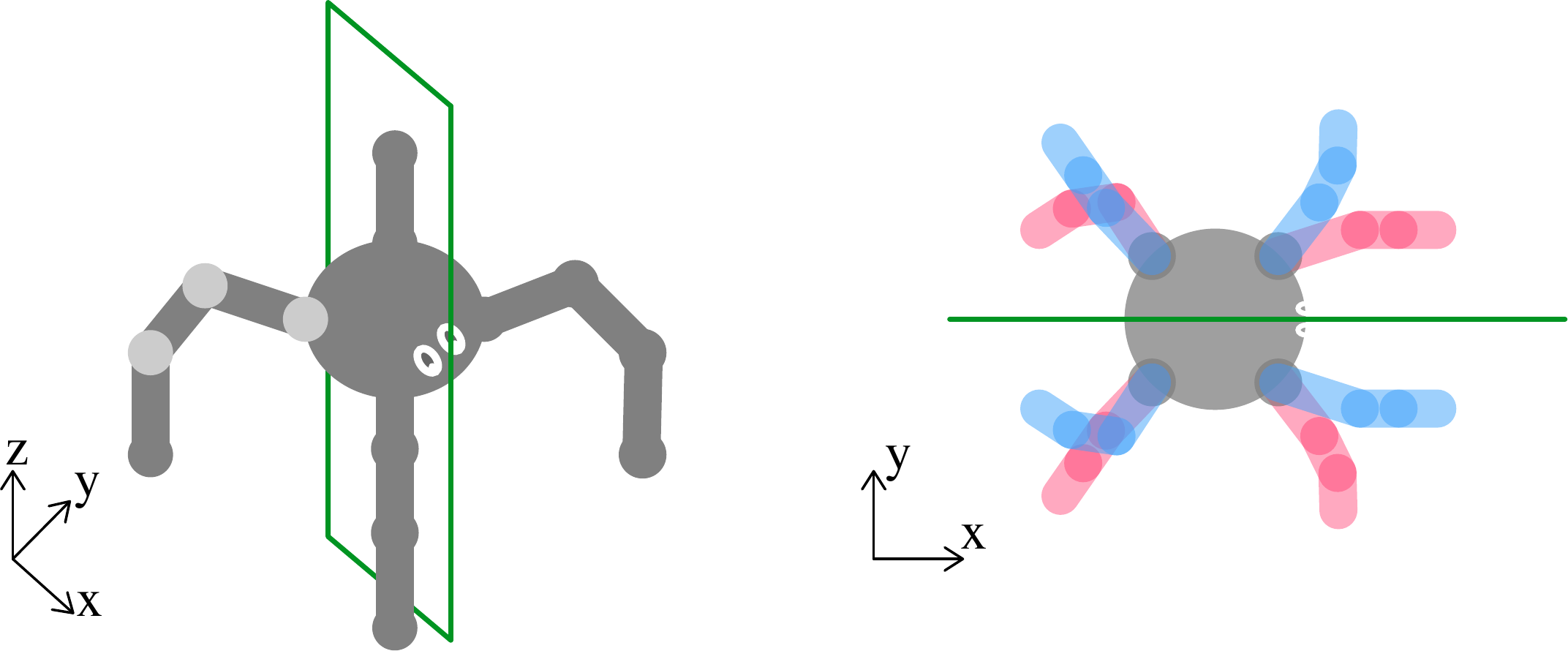}
\caption{Left: Quadruped domain, showing the three joints on one of the legs (light grey), and the sagittal plane in green. Right: Top view of a quadruped, showing a fictitious adopted pose in terms of leg orientations (blue), and its corresponding pose (pink), mirrored about the axis of symmetry. The co-ordinate axes correspond to the local frame of reference of the torso. }
\label{fig:schematic}
\end{wrapfigure}

\label{sec:Method-symm}
To illustrate the use of symmetry, we can use any animal-like domain with a saggital plane of symmetry, which is a plane of symmetry common to several animals. We use the quadruped domain from Tassa \emph{et al.} (2018) \cite{deepmindcontrolsuite2018}. The body of the quadruped comprises a torso with four legs. Each leg is identical, with three hinge joints, as shown in Figure \ref{fig:schematic} (left). The joints are controlled by torque actuators. We consider the {\emph{move}} tasks in Tassa \emph{et al.} (2018) \cite{deepmindcontrolsuite2018}, in which positive reward is given for an upright orientation relative to the horizontal plane and a forward velocity of the torso, relative to a specified desired velocity, in the frame of reference of the torso. Within the {\emph{move}} tasks, the {\emph {walk}} task and {\emph{run}} task are specified by different values of desired velocity. Details of the reward function are given in Tassa \emph{et al.} (2018) \cite{deepmindcontrolsuite2018}. The observations comprise joint angles and velocities, forces and torques on the joints, the state of the actuators, and the torso velocity. 

This is illustrated in Figure \ref{fig:schematic} (right), by means of a pose with the legs shown in blue, and its corresponding {\emph{mirrored}} pose, with the legs shown in pink. In the {\emph{move}} tasks, for every observation-action pair that the agent encounters, there is a corresponding mirrored observation-action pair that can be obtained by reflection about the plane of symmetry. The mirroring operation for observations and actions are essentially matrix multiplications, and are obtained cheaply from the observations. The mirrored corresponding observation-action pair will result in the same forward displacement, and therefore the same reward. {\bf Therefore, for every trajectory executed by the agent, there is a corresponding mirrored trajectory that results in the same sequence of rewards.}

\section{Symmetric policy} 
A \emph{symmetric gait} such as a human walk is characterised by a phase difference between the legs, and an \emph{antisymmetric gait} such as that of a galloping horse is characterised by an in-phase synchrony between the left and right halves of the body. When we use the term \emph {symmetric policy}, we mean that the probability $\pi\left(a_{\mathrm{mirrored}}, s_{\mathrm{mirrored}}\right) = \pi(a, s)$, where $s_{\mathrm{mirrored}}$ and $a_{\mathrm{mirrored}}$ are obtained by reflecting the state $s$ and action $a$, respectively, about a plane of symmetry. By this definition, the biological \emph{symmetric} and \emph{antisymmetric} gaits are both executed by a symmetric policy.  

\section{Proposed algorithm}
We use MPO, an actor-critic algorithm described in Abdolmaleki \emph{et al.}\ \cite{abdolmaleki2018maximum}. To encourage the agent to take symmetries into account, we  augment the batch of experiences generated by the \emph{actor} and stored in the {\emph{replay buffer}}, by computing a set of corresponding mirrored trajectories for use by the \emph{learner} of the MPO algorithm \cite{abdolmaleki2018maximum}. The MPO algorithm has two steps, \textbf{policy evaluation} and \textbf{policy improvement}. In this section, we summarise the algorithm and specify the modifications required to use data from mirrored trajectories, with a pseudocode of our approach given in algorithm \ref{Alg:mpo}.
\begin{algorithm}[t]
\small
\caption{MPO with Mirrored Data}\label{Alg:mpo}
\begin{algorithmic}[1]
\STATE {\bf given} batch-size (K), number of actions (N), Q-function $\hat{Q}$, old-policy $\pi_{k}$ and replay-buffer
\STATE {\bf initialize $\pi_\theta$ from the parameters of $\pi^{(k)}$} 
\REPEAT
\STATE {Sample states with batch of size N from replay buffer}
\STATE {\bf Step 1: sample based policy (weights)}
\STATE $q(a_i | s_j) = q_{ij}$, {\bf computed as:}
\FOR{j = 1,...,$K$} 
\FOR{i = 1,...,$N$}
\STATE {$a_{i} \sim \pi_\text{k}(a|s_j)$} 
\STATE {$Q_{ij} = \hat{Q}(s_{j}, a_i)$} 
\STATE $q_{ij} \propto \exp(Q_{ij}/\mathrm{temperature\  parameter})$)
\ENDFOR
\ENDFOR
\STATE {Calculate mirrored states and mirrored actions}
\STATE {\bf Step 2: update parametric policy}
\STATE {Given the data-set $\{s_j,(a_{i},q_{ij})_{i=1...N}\}_{j=1...K}$ and $\{s^{\textrm{mirrored}}_j,(a^{\textrm{mirrored}}_{i},q_{ij})_{i=1...N}\}_{j=1...K}$}
\STATE {\bf Update the Policy by taking gradient of following weighted maximum likelihood objective }
\STATE $\textrm{max}_\theta (\sum_j^K \sum_i^N q_{ij} \log \pi_{\theta}(a_i|s_j) + \sum_j^K \sum_i^N q_{ij} \log \pi_{\theta}(a^{\textrm{mirrored}}_i|s^{\textrm{mirrored}}_j) )$
\STATE {\bf (subject to additional (KL) regularization)}
\UNTIL{Fixed number of steps}
\STATE return $\pi_{(k+1)}$
\end{algorithmic}
\end{algorithm}
\subsection{Policy Evaluation}
We employ a 1-step temporal difference (TD) learning, fitting a parametric Q-function $Q_\phi^\pi(s, a)$ with parameters $\phi$ by minimizing the squared (TD) error
\begin{equation*}
\textstyle{
    \min_\phi \left(r_t + \gamma Q_{\phi'}^{\pi^{(k-1)}}\left(s_{t+1}, a_{t+1}\right) - Q_\phi^{\pi^{(k)}}\left(s_t, a_t\right)\right) ^2,
}
\end{equation*}
where $a_{t+1}\sim\pi^{(k-1)}(a|s_{t+1})$ and $r_t = r(s_t, a_t)$, which we optimize via gradient descent. We let $\phi'$ be the parameters of a target network that is held constant for $250$ steps (and then copied from the optimized parameters $\phi$). 
Using the fact that for a symmetric policy, ${Q}\left(s_\mathrm{mirrored},a_{\mathrm{mirrored}}\right) = Q(s,a)$, we also fit the parametric Q-function for ${Q}\left(s_\mathrm{mirrored},a_{\mathrm{mirrored}}\right)$,
\begin{equation*}
\textstyle{
    \min_\phi \left(r_t + \gamma Q_{\phi'}^{\pi^{(k-1)}}\left(s_{t+1}, a_{t+1}\right) - Q_\phi^{\pi^{(k)}}\left(s^\textrm{mirrored}_t, a^\textrm{mirrored}_t\right)\right) ^2.
}
\end{equation*}

\subsection{Policy Improvement}
\textbf{Step 1:} We construct a non-parametric improved policy $q$. This is done by maximizing $\bar{J}(s,q) = \mathbb{E}_{q(a|s)}[\hat{Q}(s,a)]$ for states $s$ drawn from a replay buffer $\mathcal{R}$ while ensuring that the solution stays close to the current policy $\pi_k$; i.e. $\mathbb{E}_{s \sim \mathcal{R}}[\text{KL}(q(a|s), \pi_{k}(a|s))] < \epsilon$, as detailed in Abdolmaleki \emph{et al.}\ \citep{abdolmaleki2018maximum}. \\
\textbf{Step 2:} We fit a new parametric policy to samples from $q(a|s)$ by solving the optimization problem 
${\pi_{k+1} = \mathrm{argmin}_{\pi} \mathbb{E}_{\mu(s)}[\text{KL}(q(a|s)\Vert \pi(a\vert s)]}$, where $\pi_{k+1}$ is the new and improved policy, which employs additional regularization \citep{abdolmaleki2018maximum}.
To learn about/from mirrored data, we repeat steps 1 and 2 for mirrored states and mirrored actions calculated from original data. 

\section{Experiments}

\begin{figure}[t!]
    \centering
    \includegraphics[width=0.90\columnwidth]{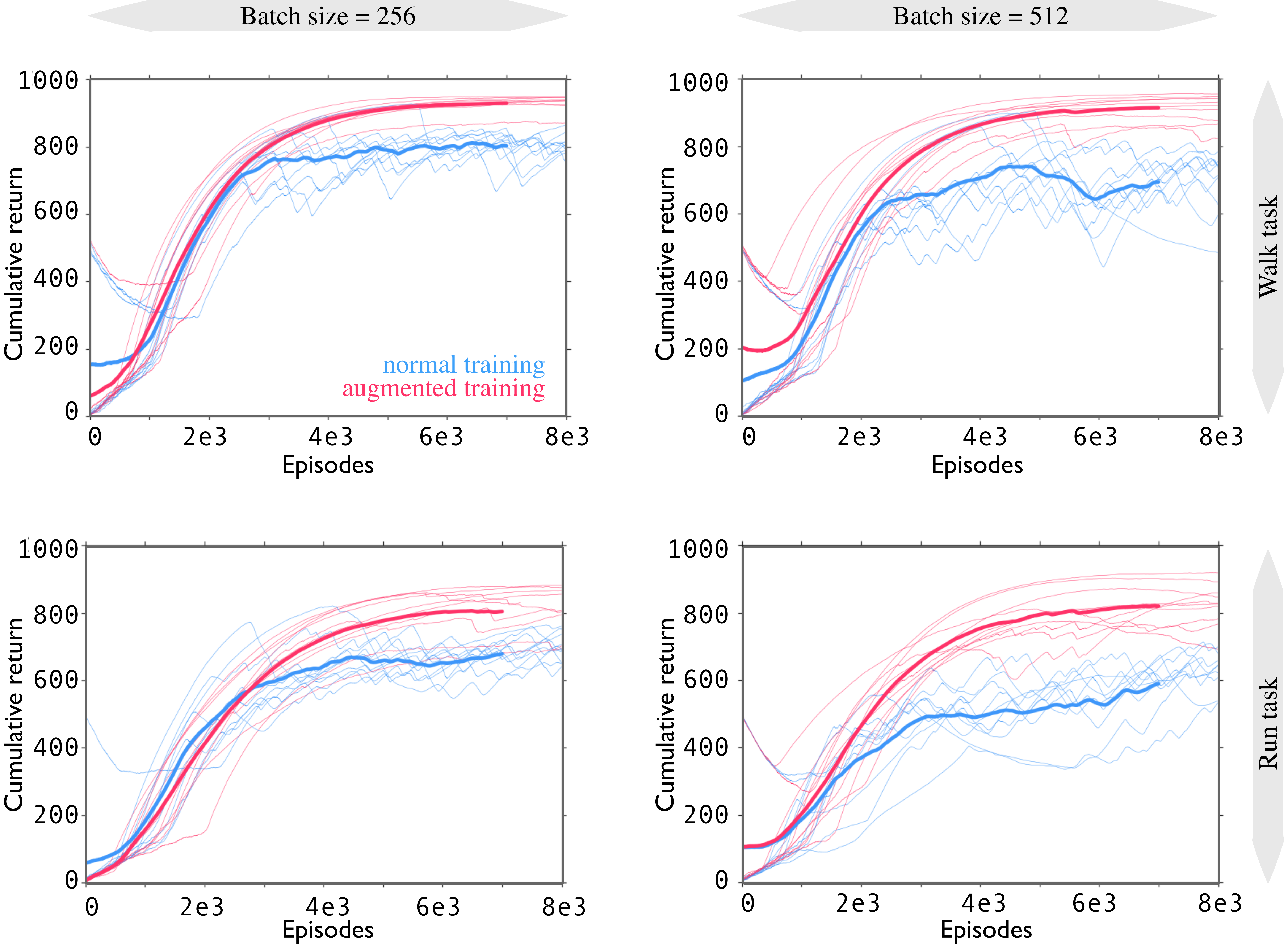}
    \caption{\textbf{Learning curves} for 10 seeds of the walk (top row) and run (bottom row) tasks, for batch sizes 256 (left column) and 512 (right column). The plot shows the cumulative reward as a function of gradient steps. The normal training and augmented training are in blue and pink, respectively. The thick line shows the mean.}
    \label{fig:exp-reward}
\end{figure}

We refer to  training with the MPO algorithm in \cite{abdolmaleki2018maximum} as ``normal training'' and training with the additional experience from the mirrored data, as described in Section \ref{sec:Method}, as ``augmented training''. For the normal and augmented training conditions, we use the same hyper-parameters as in \cite{abdolmaleki2018maximum}, with one main modification: we slowed down the actor to get a rate of trajectory generation of 1000 s of environment steps every 30 s of real time, to bring it closer to the data-generation rate of real robots. The learner thus operates in a data-limited regime. The resulting cumulative reward as a function of gradient steps is shown in Figure \ref{fig:exp-reward}. For each of the training conditions, the augmented training condition achieves better performance, typically in fewer episodes. This preliminary result suggests that knowledge of symmetry, enforced through data augmentation in the policy optimisation step, is a useful bias for the agent to shape its policy.

\section{Discussion} 
We presented an approach for incorporating symmetry in the environment in an actor-critic architecture by augmenting the experiences generated by the actor. A different approach to inform an agent of symmetry in its domain could, for instance, compress the state representation, explicitly specifying the relationship between an observation-action pair and its mirrored version. This involves changing the underlying Markov decision process so that transitions are between ``canonical" observations. In general, this means that transitions do not arise from a physical process and may require modifications to the algorithm to ensure that no undesirable bias has been introduced. 
A promising extension of this work would be to move towards a policy that explicitly encodes the symmetry using hierarchy, with similar legs sharing their lower-level policy with the corresponding actuators of other legs, and using experiences from the other legs to update their policies. Using symmetric policies leads to a change in the space of policies that can be achieved and therefore qualitative change in gait. We plan to investigate this further. In general, this work can also be useful for transfer learning to tasks which are not symmetric; an agent may first learn to walk using a knowledge of symmetry to be data-efficient, and then learn to navigate an external non-uniform landscape. 

\section{Acknowledgements} 
We would like to thank Ankit Anand, Eser Ayg{\"u}n, Tom Erez, Philippe Hamel, Yuval Tassa and Daniel Toyama. 

\bibliography{arxiv}

\begin{thebibliography}{10}

\bibitem{abdolmaleki2018maximum}
Abbas Abdolmaleki, Jost~Tobias Springenberg, Yuval Tassa, Remi Munos, Nicolas
  Heess, and Martin Riedmiller.
\newblock Maximum a posteriori policy optimisation.
\newblock {\em arXiv preprint arXiv:1806.06920}, 2018.

\bibitem{baird1992document}
Henry~S Baird.
\newblock Document image defect models.
\newblock In {\em Structured Document Image Analysis}, pages 546--556.
  Springer, 1992.

\bibitem{barenblatt1996scaling}
Grigory~Isaakovich Barenblatt.
\newblock {\em Scaling, self-similarity, and intermediate asymptotics:
  dimensional analysis and intermediate asymptotics}, volume~14.
\newblock Cambridge University Press, 1996.

\bibitem{bengio2009curriculum}
Yoshua Bengio, J{\'e}r{\^o}me Louradour, Ronan Collobert, and Jason Weston.
\newblock Curriculum learning.
\newblock In {\em Proceedings of the 26th annual international conference on
  machine learning}, pages 41--48. ACM, 2009.

\bibitem{bohez2019value}
Steven Bohez, Abbas Abdolmaleki, Michael Neunert, Jonas Buchli, Nicolas Heess,
  and Raia Hadsell.
\newblock Value constrained model-free continuous control.
\newblock {\em arXiv preprint arXiv:1902.04623}, 2019.

\bibitem{haarnoja2018soft}
Tuomas Haarnoja, Aurick Zhou, Kristian Hartikainen, George Tucker, Sehoon Ha,
  Jie Tan, Vikash Kumar, Henry Zhu, Abhishek Gupta, Pieter Abbeel, et~al.
\newblock Soft actor-critic algorithms and applications.
\newblock {\em arXiv preprint arXiv:1812.05905}, 2018.

\bibitem{krizhevsky2012imagenet}
Alex Krizhevsky, Ilya Sutskever, and Geoffrey~E Hinton.
\newblock Imagenet classification with deep convolutional neural networks.
\newblock In {\em Advances in neural information processing systems}, pages
  1097--1105, 2012.

\bibitem{lillicrap2015continuous}
Timothy~P Lillicrap, Jonathan~J Hunt, Alexander Pritzel, Nicolas Heess, Tom
  Erez, Yuval Tassa, David Silver, and Daan Wierstra.
\newblock Continuous control with deep reinforcement learning.
\newblock {\em International Conference on Learning Representations}, 2016.

\bibitem{merel2018neural}
Josh Merel, Leonard Hasenclever, Alexandre Galashov, Arun Ahuja, Vu~Pham, Greg
  Wayne, Yee~Whye Teh, and Nicolas Heess.
\newblock Neural probabilistic motor primitives for humanoid control.
\newblock {\em arXiv preprint arXiv:1811.11711}, 2018.

\bibitem{silver2016mastering}
David Silver, Aja Huang, Chris~J Maddison, Arthur Guez, Laurent Sifre, George
  Van Den~Driessche, Julian Schrittwieser, Ioannis Antonoglou, Veda
  Panneershelvam, Marc Lanctot, et~al.
\newblock Mastering the game of go with deep neural networks and tree search.
\newblock {\em nature}, 529(7587):484, 2016.

\bibitem{simard2003best}
Patrice~Y Simard, David Steinkraus, John~C Platt, et~al.
\newblock Best practices for convolutional neural networks applied to visual
  document analysis.
\newblock In {\em Icdar}, volume~3, 2003.

\bibitem{deepmindcontrolsuite2018}
Yuval Tassa, Yotam Doron, Alistair Muldal, Tom Erez, Yazhe Li, Diego
  de~Las~Casas, David Budden, Abbas Abdolmaleki, Josh Merel, Andrew Lefrancq,
  Timothy Lillicrap, and Martin Riedmiller.
\newblock Deep{Mind} control suite.
\newblock https://arxiv.org/abs/1801.00690, January 2018.

\bibitem{wilson1971renormalization}
Kenneth~G Wilson.
\newblock Renormalization group and critical phenomena. i. renormalization
  group and the kadanoff scaling picture.
\newblock {\em Physical review B}, 4(9):3174, 1971.

\end{thebibliography}
\bibliographystyle{plain}

\end{document}